\documentclass{article}

\PassOptionsToPackage{numbers, compress}{natbib}

\usepackage[final]{neurips_wrl2023}

\usepackage[utf8]{inputenc} 
\usepackage[T1]{fontenc}    
\usepackage{hyperref}       
\usepackage{url}            
\usepackage{booktabs}       
\usepackage{amsfonts}       
\usepackage{nicefrac}       
\usepackage{microtype}      

\usepackage{graphicx} 
\usepackage{float}
\usepackage{amsmath}
\usepackage{wrapfig}
\usepackage{siunitx}

\usepackage{caption}
\usepackage{subcaption}

\usepackage{pgfplots}
\DeclareUnicodeCharacter{2212}{−}
\usepgfplotslibrary{groupplots,dateplot}
\usetikzlibrary{patterns,shapes.arrows}
\pgfplotsset{compat=newest}

\usepackage{authblk}

\title{Swarm-GPT: Combining Large Language Models with Safe Motion Planning for Robot Choreography Design}

\author[1]{Aoran Jiao}
\author[1]{Tanmay P. Patel}
\author[1]{Sanjmi Khurana}
\author[1]{Anna-Mariya Korol}
\author[2,3]{Lukas Brunke}
\author[3]{\newline Vivek K. Adajania}
\author[2]{Utku Culha}
\author[2]{Siqi Zhou}
\author[2,3]{Angela P. Schoellig}

\affil[1]{Department of Engineering Science, University of Toronto}
\affil[2]{Department of Computer Engineering, Technical University of Munich}
\affil[3]{Institute for Aerospace Studies, University of Toronto}

\begin{document}
\maketitle
\begin{abstract}

  
This paper presents Swarm-GPT, a system that integrates large language models~(LLMs) with safe swarm motion planning--- offering an automated and novel approach to deployable drone swarm choreography. Swarm-GPT enables users to automatically generate synchronized drone performances through natural language instructions. With an emphasis on safety and creativity, Swarm-GPT addresses a critical gap in the field of drone choreography by integrating the creative power of generative models with the effectiveness and safety of model-based planning algorithms. This goal is achieved by prompting the LLM to generate a unique set of waypoints based on extracted audio data. A trajectory planner processes these waypoints to guarantee collision-free and feasible motion. Results can be viewed in simulation prior to execution and modified through dynamic re-prompting. Sim-to-real transfer experiments demonstrate Swarm-GPT's ability to accurately replicate simulated drone trajectories, with a mean sim-to-real root mean square error (RMSE) of 28.7 mm. To date, Swarm-GPT has been successfully showcased at three live events, exemplifying safe real-world deployment of pre-trained models.
\end{abstract}

\section{Introduction}
For millennia, dance has been a powerful medium for human expression and entertainment. However, our mastery of robot dynamics and control has only recently expanded to the realm of dancing and performance. 
One illustrative application of robot choreography in the entertainment industry is drone shows, where many unmanned aerial vehicles (UAVs) morph into intricate patterns in mid-air, synchronized with musical rhythms~\cite{sparked,schollig2010synchronizing,du2019fast}.

Imparting dance to robots was a natural progression but yet non-trivial~\cite{apostolos1996robot}. In current applications, the robots' spatiotemporal movements are often manually choreographed by experts to balance expressivity and safety. Generating robot choreographies can be laborious; as the number of robots increases, the complexity in choreographic design and safety analysis can quickly become intractable~\cite{waibel_drone_2017}.

In this work, we propose Swarm-GPT --- an automated drone choreography pipeline that translates language instructions into choreographed motions of a swarm of nano-drones. To facilitate intuitive interaction and safe deployment of language-based drone choreography, the system combines \textit{(a)} a high-level motion generation layer that leverages the generative abilities of the large language model (LLM) to design unique drone swarm choreographies and \textit{(b)} a low-level safety layer that uses model-based swarm planning frameworks to ensure feasible and safe deployment of the LLM-generated choreographies to physical robots. 

Through simulation and physical experiments, we show how to effectively bridge the gap between high-level natural language instructions and low-level robot control and coordination. Swarm-GPT is a proof-of-concept showcasing the viability of LLM-directed drone swarm control, which can be accomplished safely. Our work demonstrates that LLMs can be used as an intuitive interface for non-expert users to generate complex drone behaviours while being augmented by underlying safe control and planning algorithms to enable safe coordination by design.

To the best of our knowledge, Swarm-GPT is the first system that enables the direct use of LLMs for drone choreography. Our contributions are as follows: \textit{(i)} enabling non-experts to program and modify choreographies of drone swarms by using natural language, \textit{(ii)} seamlessly integrating large language models with a model-based safety filter to guarantee safe execution, and \textit{(iii)} demonstrating Swarm-GPT in real-time drone experiments.
\vspace{-0.3cm}


\begin{figure}
    \centering
    \includegraphics[width=\textwidth]{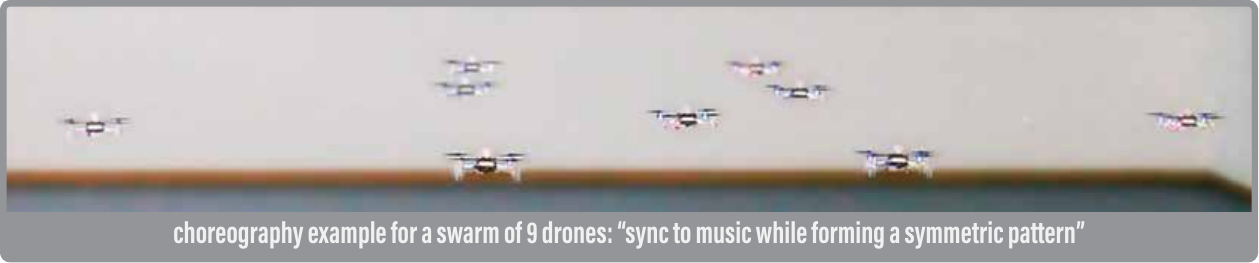}
    \vspace{-10pt}
    \caption{Choreography of a 9-drone swarm by Swarm-GPT. Formations shown in the picture are synchronized to the music beat times. Users can interact with Swarm-GPT to generate different trajectories. Full experimental
performance available from the video: \href{http://tiny.cc/Swarm-GPT}{http://tiny.cc/Swarm-GPT}}
    \vspace{-15pt}
    \label{fig:a}
\end{figure}

\section{Related Work}

\textbf{Robot Choreography.}
The study of robot choreography spans across different robot platforms, including both humanoid robots~\cite{boston_dynamics,guizzo2019leaps,xia2012autonomous, boukheddimi2022robot} and non-anthropomorphic forms such as robot arms~\cite{rogel2022robogroove}, drones~\cite{sparked,schollig2010synchronizing,du2019fast}, and quadruped robots~\cite{boston_dynamics,guizzo2019leaps}. While previous efforts have focused on automating the process by mapping music features(as in~\cite{guizzo2019leaps,xia2012autonomous,schollig2010synchronizing,du2019fast}), the choreography process still often requires manual tuning and domain expertise, and provides minimal means for intuitive feedback, especially for non-expert users. This work focuses on drone swarm choreographies and introduces an automated process with a natural language feedback tool for synchronized spatiotemporal motion generation.

\textbf{LLMs for Robotics.} Robot decision-making through natural language recently started gaining traction~\cite{bommasani2021opportunities,vemprala_chatgpt_2023}, especially in the areas of visual-language navigation~\cite{huang2023visual}, language-grounded robot affordances~\cite{brohan2023can}, language-based trajectory modification~\cite{bucker2023latte}, and high-level task planning through LLM-based code-generation~\cite{singh2023progprompt, tang_saytap:_2023}. 
These examples highlight the potential of leveraging LLMs to facilitate natural interaction with robotic systems. In this work, we pursue a similar goal but with a focus on language-based drone swarm choreography and its safe real-world deployment. 

\textbf{Safe Robot Swarm Decision-Making.}
Safe robot decision-making approaches generally exploit our prior knowledge about the robot to provide desired safety guarantees~\cite{brunke2022safe}. In the context of swarm coordination, model-based motion planning offers a natural means to explicitly account for safety and feasibility constraints. Recent advances in model-based swarm motion planning include both centralized methods~\cite{augugliaro2012generation, mip_how} and distributed methods~\cite{luis-ral20,soria2021distributed,adajania_amswarm:_2023}. Given the scalability advantages of distributed methods~\cite{luis-ral20,soria2021distributed,adajania_amswarm:_2023}, in this work, we adopt a state-of-the-art distributed drone swarm motion planning framework~\cite{adajania_amswarm:_2023} as a safety filter for LLM-based swarm choreography design. This integration addresses the challenge of safe deployment by effectively leveraging our prior knowledge about the robots while maximally preserving the motion generated by the LLM.

\section{Methodology}
\label{sec: Swarm-GPT_design}
In this section, we provide an overview of the proposed Swarm-GPT framework. A block diagram of the Swarm-GPT system is shown in Figure~\ref{fig:block_diagram} and further technical details are included in Appendix~\ref{appendix: more details on methodology}.

\textbf{LLM Interface.}
The LLM interface offers a means for users to provide high-level task specifications such as song selection and swarm behaviour through natural language. In parallel, the features of the selected song are extracted using an audio analysis tool, librosa,~\cite{mcfee_librosa/librosa:_2023}. The language input and extracted music features together formulate an LLM prompt template with preliminary instructions about the physical constraints of the operating environment. We include details of the prompting process in Appendix~\ref{appendix: more details on llm interface}. Given the language input, song audio file, and high-level prior knowledge of the environment, the LLM generates choreography for the drone swarm as a series of timed position waypoints for each drone, synchronized with the beats of the selected song. 

\begin{wrapfigure}{l}{0.55\textwidth}
    \centering
    \vspace{-10pt}
    \includegraphics[width=0.55\columnwidth]{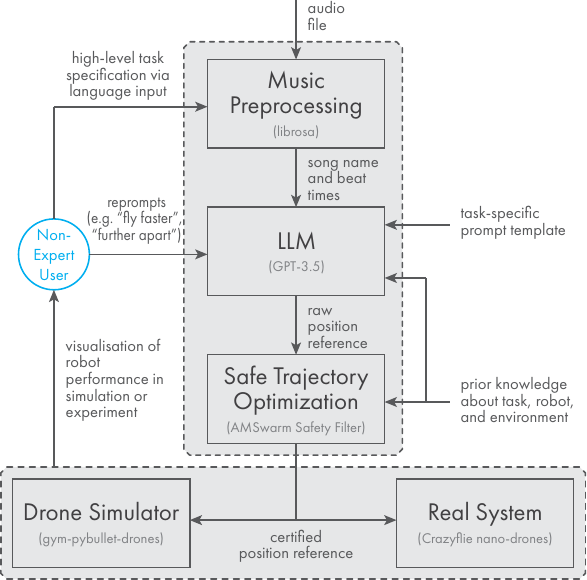}
    \caption{Swarm-GPT system block diagram}
    \vspace{-10pt}
    \label{fig:block_diagram}
\end{wrapfigure}

\textbf{Safe Trajectory Optimization.} 
The safe trajectory optimization module is a safety filter formulated based on a state-of-the-art distributed drone swarm motion planning framework, the AMSwarm~ \cite{adajania_amswarm:_2023}.
The AMSwarm safety filter allows us to incorporate our prior knowledge about the robot system (e.g., maximum allowable speed and actuation limits of the drones as well as ellipsoidal collision envelopes of the drones) and subsequently computes feasible and collision-free swarm trajectories that best match the choreography waypoints generated by the LLM. 
The formulation of the AMSwarm-based safety filter design is further discussed in Appendix~\ref{methodology:collision-avoidance}.


\textbf{Trajectory Modification (Re-Prompting).}
These waypoints are treated as position commands and are visualized using a drone swarm simulator, the gym-pybullet-drones~\cite{panerati2021learning}. 
The user may deploy the generated choreography onto the real drone system or modify it by re-prompting. To re-prompt, the user can enter a custom prompt (e.g., ``go faster, higher''), which is appended to the original prompt. The LLM then modifies the swarm by updating the waypoints based on the reprompt. 
Collision avoidance and processing are applied to the updated LLM outputs.


\section{Simulation and Experimental Evaluation}
\label{section:results}
The proposed Swarm-GPT system was tested with four songs of different genres \{\emph{Perfect}, \emph{Ode to Joy}, \emph{Mission Impossible Theme}, \emph{Here Comes the Sun}\} and swarms up to nine drones in both simulation and real-time experiments. In this section, we present key evaluation results; additional discussions are included in Appendix~\ref{appendix:results}. A video of the robot experiments is available at this link: \href{http://tiny.cc/Swarm-GPT}{http://tiny.cc/Swarm-GPT}.

\textbf{Assessment of Sim-to-Real Transfer.} We use gym-pybullet-drones \cite{panerati2021learning} as the simulator of Swarm-GPT. The simulator is equipped with the same controllers as the real drone platform. To examine the sim-to-real transfer gap, the control signal input into the simulator and the real system are identical. The root mean square error (RMSE) of the simulated $x$, $y$, $z$ positions and the ground truth $x$, $y$, $z$ positions are plotted in the bar graph for six different choreography configurations. As Fig.~\ref{fig:p} illustrates, the average sim-to-real RMSE among the swarm of drones does not exceed 60 mm in the worst case. The average RMSE across the six choreographies is 28.7 mm. With the high sim-to-real transfer quality, we use simulation to collect larger quantities of data to illustrate the safety features and generative abilities of Swarm-GPT.

\textbf{Response to Safety Constraints.} LLMs are excellent creative tools, but they have limited awareness of dynamics and robot constraints. Our pipeline complements the creative power of LLMs with well-tested and robust safety filters. Collectively, the pipeline achieves safe drone motion that humans can design through natural language. Fig.~\ref{fig:5} shows the number of collision events before and after applying our safety filter. To count collision events, we consider ellipsoidal collision-avoidance envelopes that account for both true dimensions and aerodynamic effects (e.g., downwash) of the drones~\cite{adajania_amswarm:_2023}. Across 97 simulation trials, only $21.65\%$ of the raw LLM-generated trajectories were collision-free. This figure reached $100\%$ after the safety algorithm, meaning all collisions were eliminated.  We note that, during real-world deployment, there could be additional dynamic uncertainties or imperfections; these uncertainties could be systematically accounted for by using robustly safe formulations~\cite{brunke2022safe}.

\begin{figure}[ht]
    \begin{minipage}[c]{0.5\columnwidth}
    \hspace{-4.5em}
        \centering
\begin{tikzpicture}

\definecolor{crimson2143940}{RGB}{214,39,40}
\definecolor{darkgray176}{RGB}{176,176,176}
\definecolor{forestgreen4416044}{RGB}{44,160,44}
\definecolor{lightgray204}{RGB}{204,204,204}
\definecolor{steelblue31119180}{RGB}{31,119,180}

\pgfplotsset{compat=1.11,
    /pgfplots/ybar legend/.style={
    /pgfplots/legend image code/.code={%
       \draw[##1,/tikz/.cd,yshift=-0.25em]
        (0cm,0cm) rectangle (3pt,0.8em);},
   },
}

\begin{axis}[
legend cell align={left},
legend style={fill opacity=0.8, draw opacity=1, text opacity=1, draw=lightgray204},
tick align=outside,
tick pos=left,
x grid style={darkgray176},
xmin=-0.8, xmax=6.2,
xtick style={color=black},
xticklabel style={rotate=45.0,anchor=east},
xtick={0, 1, 2, 3, 4, 5},
xticklabels={C1, C2, C3, C4, C5, C6},
ytick={0.0, 0.02, 0.04, 0.06},
yticklabels={0, 20, 40, 60},
y grid style={darkgray176},
ylabel={RMSE (mm)},
xlabel={Choreography Index},
ymin=0, ymax=0.0788131309245,
ytick style={color=black},
scaled y ticks=false,
width=6cm,
height=4cm,
]


\draw[draw=black,fill=forestgreen4416044] (axis cs:-0.4,0) rectangle (axis cs:0.4,0.01784430776);
\addlegendimage{ybar,ybar legend,draw=black,fill=crimson2143940}

\draw[draw=black,fill=forestgreen4416044] (axis cs:0.6,0) rectangle (axis cs:1.4,0.02980021197);

\draw[draw=black,fill=forestgreen4416044] (axis cs:1.6,0) rectangle (axis cs:2.4,0.05523411821);
\addlegendimage{ybar, ybar legend,draw=black,fill=steelblue31119180}

\draw[draw=black,fill=forestgreen4416044] (axis cs:2.6,0) rectangle (axis cs:3.4,0.0329562476);

\draw[draw=black,fill=forestgreen4416044] (axis cs:3.6,0) rectangle (axis cs:4.4,0.01999381102);
\draw[draw=black,fill=forestgreen4416044] (axis cs:4.6,0) rectangle (axis cs:5.4,0.02137486751);
\addlegendimage{ybar,ybar legend,draw=black,fill=forestgreen4416044}

\path [draw=black, semithick]
(axis cs:0,0.01678430776)
--(axis cs:0,0.01890430776);

\path [draw=black, semithick]
(axis cs:1,0.027090327723)
--(axis cs:1,0.032510096217);

\path [draw=black, semithick]
(axis cs:2,0.03540811173)
--(axis cs:2,0.07506012469);

\path [draw=black, semithick]
(axis cs:3,0.0225562476)
--(axis cs:3,0.0433562476);

\path [draw=black, semithick]
(axis cs:4,0.01703381102)
--(axis cs:4,0.02295381102);

\path [draw=black, semithick]
(axis cs:5,0.01553486751)
--(axis cs:5,0.02721486751);

\addplot [semithick, black, mark=-, mark size=5, mark options={solid}, only marks, forget plot]
table {%
0 0.01678430776
1 0.027090327723
2 0.03540811173
3 0.0225562476
4 0.01703381102
5 0.01553486751
};
\addplot [semithick, black, mark=-, mark size=5, mark options={solid}, only marks, forget plot]
table {%
0 0.01890430776
1 0.032510096217
2 0.07506012469
3 0.0433562476
4 0.02295381102
5 0.02721486751
};
\addplot [semithick, black, dashed, forget plot]
table {%
-1 0.029533927345
6 0.029533927345
};
\draw (axis cs:3.5,0.030533927345) node[
  scale=0.5,
  anchor=base west,
  text=black,
  rotate=0.0
]{RMSE};
\end{axis}

\end{tikzpicture}
        \caption{Summary of sim-to-real RMSE for\\ six different choreographies.}
        \label{fig:p}
    \end{minipage}
    \begin{minipage}[c]{0.5\columnwidth}
        \centering
\begin{tikzpicture}

\definecolor{darkgray176}{RGB}{176,176,176}
\definecolor{darkorange25512714}{RGB}{255,127,14}
\definecolor{lightgray204}{RGB}{204,204,204}
\definecolor{steelblue31119180}{RGB}{31,119,180}

\begin{axis}[
height=4cm,
legend cell align={left},
legend style={fill opacity=0.8, draw opacity=1, text opacity=1, draw=lightgray204},
tick align=outside,
tick pos=left,
width=6cm,
x grid style={darkgray176},
xlabel={Percent of Trajectory with Collisions [\%]},
xmin=-4.5, xmax=94.5,
xtick style={color=black},
y grid style={darkgray176},
ylabel={Frequency},
ymin=0, ymax=100.8,
ytick style={color=black}
]
\draw[draw=none,fill=steelblue31119180,fill opacity=0.8] (axis cs:0,0) rectangle (axis cs:5,96);
\addlegendimage{ybar,ybar legend,draw=none,fill=steelblue31119180,fill opacity=0.8}
\addlegendentry{After Filtering}

\draw[draw=none,fill=steelblue31119180,fill opacity=0.8] (axis cs:5,0) rectangle (axis cs:10,1);
\draw[draw=none,fill=steelblue31119180,fill opacity=0.8] (axis cs:10,0) rectangle (axis cs:15,0);
\draw[draw=none,fill=steelblue31119180,fill opacity=0.8] (axis cs:15,0) rectangle (axis cs:20,0);
\draw[draw=none,fill=steelblue31119180,fill opacity=0.8] (axis cs:20,0) rectangle (axis cs:25,0);
\draw[draw=none,fill=steelblue31119180,fill opacity=0.8] (axis cs:25,0) rectangle (axis cs:30,0);
\draw[draw=none,fill=steelblue31119180,fill opacity=0.8] (axis cs:30,0) rectangle (axis cs:35,0);
\draw[draw=none,fill=steelblue31119180,fill opacity=0.8] (axis cs:35,0) rectangle (axis cs:40,0);
\draw[draw=none,fill=steelblue31119180,fill opacity=0.8] (axis cs:40,0) rectangle (axis cs:45,0);
\draw[draw=none,fill=steelblue31119180,fill opacity=0.8] (axis cs:45,0) rectangle (axis cs:50,0);
\draw[draw=none,fill=steelblue31119180,fill opacity=0.8] (axis cs:50,0) rectangle (axis cs:55,0);
\draw[draw=none,fill=steelblue31119180,fill opacity=0.8] (axis cs:55,0) rectangle (axis cs:60,0);
\draw[draw=none,fill=steelblue31119180,fill opacity=0.8] (axis cs:60,0) rectangle (axis cs:65,0);
\draw[draw=none,fill=steelblue31119180,fill opacity=0.8] (axis cs:65,0) rectangle (axis cs:70,0);
\draw[draw=none,fill=steelblue31119180,fill opacity=0.8] (axis cs:70,0) rectangle (axis cs:75,0);
\draw[draw=none,fill=steelblue31119180,fill opacity=0.8] (axis cs:75,0) rectangle (axis cs:80,0);
\draw[draw=none,fill=steelblue31119180,fill opacity=0.8] (axis cs:80,0) rectangle (axis cs:85,0);
\draw[draw=none,fill=steelblue31119180,fill opacity=0.8] (axis cs:85,0) rectangle (axis cs:90,0);
\draw[draw=none,fill=darkorange25512714,fill opacity=0.8] (axis cs:0,0) rectangle (axis cs:5,25);
\addlegendimage{ybar,ybar legend,draw=none,fill=darkorange25512714,fill opacity=0.8}
\addlegendentry{Before Filtering}

\draw[draw=none,fill=darkorange25512714,fill opacity=0.8] (axis cs:5,0) rectangle (axis cs:10,2);
\draw[draw=none,fill=darkorange25512714,fill opacity=0.8] (axis cs:10,0) rectangle (axis cs:15,4);
\draw[draw=none,fill=darkorange25512714,fill opacity=0.8] (axis cs:15,0) rectangle (axis cs:20,1);
\draw[draw=none,fill=darkorange25512714,fill opacity=0.8] (axis cs:20,0) rectangle (axis cs:25,0);
\draw[draw=none,fill=darkorange25512714,fill opacity=0.8] (axis cs:25,0) rectangle (axis cs:30,4);
\draw[draw=none,fill=darkorange25512714,fill opacity=0.8] (axis cs:30,0) rectangle (axis cs:35,1);
\draw[draw=none,fill=darkorange25512714,fill opacity=0.8] (axis cs:35,0) rectangle (axis cs:40,4);
\draw[draw=none,fill=darkorange25512714,fill opacity=0.8] (axis cs:40,0) rectangle (axis cs:45,2);
\draw[draw=none,fill=darkorange25512714,fill opacity=0.8] (axis cs:45,0) rectangle (axis cs:50,3);
\draw[draw=none,fill=darkorange25512714,fill opacity=0.8] (axis cs:50,0) rectangle (axis cs:55,3);
\draw[draw=none,fill=darkorange25512714,fill opacity=0.8] (axis cs:55,0) rectangle (axis cs:60,3);
\draw[draw=none,fill=darkorange25512714,fill opacity=0.8] (axis cs:60,0) rectangle (axis cs:65,9);
\draw[draw=none,fill=darkorange25512714,fill opacity=0.8] (axis cs:65,0) rectangle (axis cs:70,5);
\draw[draw=none,fill=darkorange25512714,fill opacity=0.8] (axis cs:70,0) rectangle (axis cs:75,7);
\draw[draw=none,fill=darkorange25512714,fill opacity=0.8] (axis cs:75,0) rectangle (axis cs:80,7);
\draw[draw=none,fill=darkorange25512714,fill opacity=0.8] (axis cs:80,0) rectangle (axis cs:85,6);
\draw[draw=none,fill=darkorange25512714,fill opacity=0.8] (axis cs:85,0) rectangle (axis cs:90,2);
\draw[draw=black,semithick] (axis cs:0,0) rectangle (axis cs:5,96);
\draw[draw=black,semithick] (axis cs:5,0) rectangle (axis cs:10,1);
\draw[draw=black,semithick] (axis cs:10,0) rectangle (axis cs:15,0);
\draw[draw=black,semithick] (axis cs:15,0) rectangle (axis cs:20,0);
\draw[draw=black,semithick] (axis cs:20,0) rectangle (axis cs:25,0);
\draw[draw=black,semithick] (axis cs:25,0) rectangle (axis cs:30,0);
\draw[draw=black,semithick] (axis cs:30,0) rectangle (axis cs:35,0);
\draw[draw=black,semithick] (axis cs:35,0) rectangle (axis cs:40,0);
\draw[draw=black,semithick] (axis cs:40,0) rectangle (axis cs:45,0);
\draw[draw=black,semithick] (axis cs:45,0) rectangle (axis cs:50,0);
\draw[draw=black,semithick] (axis cs:50,0) rectangle (axis cs:55,0);
\draw[draw=black,semithick] (axis cs:55,0) rectangle (axis cs:60,0);
\draw[draw=black,semithick] (axis cs:60,0) rectangle (axis cs:65,0);
\draw[draw=black,semithick] (axis cs:65,0) rectangle (axis cs:70,0);
\draw[draw=black,semithick] (axis cs:70,0) rectangle (axis cs:75,0);
\draw[draw=black,semithick] (axis cs:75,0) rectangle (axis cs:80,0);
\draw[draw=black,semithick] (axis cs:80,0) rectangle (axis cs:85,0);
\draw[draw=black,semithick] (axis cs:85,0) rectangle (axis cs:90,0);
\draw[draw=black,semithick] (axis cs:0,0) rectangle (axis cs:5,25);
\draw[draw=black,semithick] (axis cs:5,0) rectangle (axis cs:10,2);
\draw[draw=black,semithick] (axis cs:10,0) rectangle (axis cs:15,4);
\draw[draw=black,semithick] (axis cs:15,0) rectangle (axis cs:20,1);
\draw[draw=black,semithick] (axis cs:20,0) rectangle (axis cs:25,0);
\draw[draw=black,semithick] (axis cs:25,0) rectangle (axis cs:30,4);
\draw[draw=black,semithick] (axis cs:30,0) rectangle (axis cs:35,1);
\draw[draw=black,semithick] (axis cs:35,0) rectangle (axis cs:40,4);
\draw[draw=black,semithick] (axis cs:40,0) rectangle (axis cs:45,2);
\draw[draw=black,semithick] (axis cs:45,0) rectangle (axis cs:50,3);
\draw[draw=black,semithick] (axis cs:50,0) rectangle (axis cs:55,3);
\draw[draw=black,semithick] (axis cs:55,0) rectangle (axis cs:60,3);
\draw[draw=black,semithick] (axis cs:60,0) rectangle (axis cs:65,9);
\draw[draw=black,semithick] (axis cs:65,0) rectangle (axis cs:70,5);
\draw[draw=black,semithick] (axis cs:70,0) rectangle (axis cs:75,7);
\draw[draw=black,semithick] (axis cs:75,0) rectangle (axis cs:80,7);
\draw[draw=black,semithick] (axis cs:80,0) rectangle (axis cs:85,6);
\draw[draw=black,semithick] (axis cs:85,0) rectangle (axis cs:90,2);
\end{axis}

\end{tikzpicture}
        \caption{Histogram of the percentages of collisions along trajectories before and after applying our AMSwarm safety filter. All data points were collected with nine-drone swarms.} 
        \label{fig:5}
    \end{minipage}
\end{figure}

\begin{figure}[ht]
    \begin{minipage}[c]{0.49\columnwidth}
        \centering
        \input{figures/figure_6_2d}
        \caption{Visualization of Swarm-GPT's choreography. Different colours correspond to different drones. The large dots represent beat times, and the colour gradient moves from lighter to more saturated as time progresses.}
        \label{fig:vis}
    \end{minipage}~~
    \begin{minipage}[c]{0.49\columnwidth}
        \centering
\begin{tikzpicture}

\definecolor{darkgray176}{RGB}{176,176,176}
\definecolor{gray}{RGB}{128,128,128}
\definecolor{green01270}{RGB}{0,127,0}
\definecolor{lightgray204}{RGB}{204,204,204}

\begin{axis}[
height=4cm,
legend cell align={left},
legend style={fill opacity=0.8, draw opacity=1, text opacity=1, draw=lightgray204},
tick align=outside,
tick pos=left,
width=6cm,
x grid style={darkgray176},
xlabel={Choreography Index},
xmin=-0.35, xmax=4.6,
xtick style={color=black},
xtick={0,1,2,3,4},
xticklabel style={rotate=35.0},
xticklabels={C1,C2,C3,C4,C5},
y grid style={darkgray176},
ylabel={Average velocities (m/s)},
ymin=0, ymax=0.53698437303,
ytick style={color=black}
]
\draw[draw=none,fill=red] (axis cs:-0.125,0) rectangle (axis cs:0.125,0.2452721569);
\addlegendimage{ybar,ybar legend,draw=none,fill=red}
\addlegendentry{Before reprompting}

\draw[draw=none,fill=red] (axis cs:0.875,0) rectangle (axis cs:1.125,0.08869287518);
\draw[draw=none,fill=red] (axis cs:1.875,0) rectangle (axis cs:2.125,0.3941756148);
\draw[draw=none,fill=red] (axis cs:2.875,0) rectangle (axis cs:3.125,0.3224651201);
\draw[draw=none,fill=red] (axis cs:3.875,0) rectangle (axis cs:4.125,0.1535732826);
\draw[draw=gray,fill=green01270] (axis cs:0.125,0) rectangle (axis cs:0.375,0.3102561074);
\addlegendimage{ybar,ybar legend,draw=gray,fill=green01270}
\addlegendentry{After reprompting}

\draw[draw=gray,fill=green01270] (axis cs:1.125,0) rectangle (axis cs:1.375,0.1702602714);
\draw[draw=gray,fill=green01270] (axis cs:2.125,0) rectangle (axis cs:2.375,0.5114136886);
\draw[draw=gray,fill=green01270] (axis cs:3.125,0) rectangle (axis cs:3.375,0.4326358552);
\draw[draw=gray,fill=green01270] (axis cs:4.125,0) rectangle (axis cs:4.375,0.2311373448);
\end{axis}

\end{tikzpicture}
        \caption{The LLM can be reprompted to modify trajectories. This figure shows the mean drone velocity before and after reprompting averaged across 10 choreographies per song. The reprompt used instructed the drones to ``fly faster.''} 
        \label{fig:vel}
    \end{minipage}
\end{figure}

        \label{fig:f}






\textbf{Prompting and Synchronization to Music.}  
To ensure an accurate synchronization between the choreography and the music and facilitate intuitive interactions, we instruct ChatGPT to generate waypoints at the extracted timestamps of the beats for the music provided while achieving desired behaviours (e.g., ``symmetric formation''). An example is shown in Fig.~\ref{fig:vis}, where the ($x$, $y$) positions of 6 drones in one choreography are plotted against the beat times at each large solid dot. The fleet of drones maintains a desired formation and distinctly changes their trajectory around the beat times. Fig.~\ref{fig:vel} shows further examples of reprompting, where the drones are instructed to ``fly faster.''

\section{Conclusion}
We present a novel generative choreography system, Swarm-GPT, that integrates high-level LLMs and low-level safe drone swarm motion planning for safe and interactive drone swarm choreography. The design methods are demonstrated in experiments, showing the efficacy of combining LLM with model-based safety filter designs for safe real-world deployment.

\clearpage
\bibliographystyle{IEEEtran}
\bibliography{IEEEabrv,refs}

\clearpage
\appendix
\section{Further Details on Methodology}
\label{appendix: more details on methodology}
\subsection{LLM Interface}
\label{appendix: more details on llm interface}
\begin{figure}[hbt!]
    \centering
    \includegraphics[width=1\linewidth]{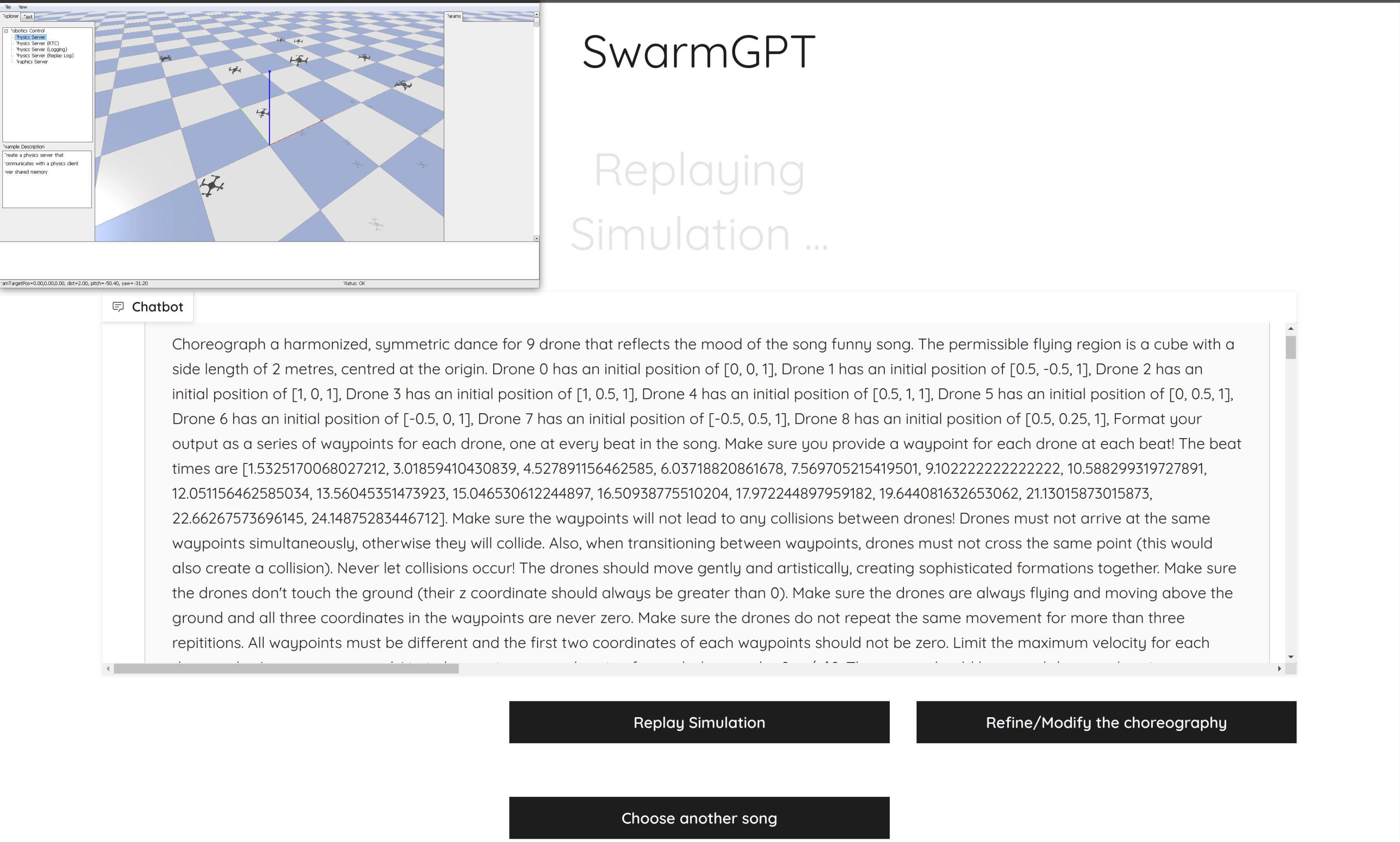}
    \vspace{0.5em}
    \caption{Illustration of the Swarm-GPT Graphical User Interface. The user starts by selecting a song from a given list. Then, the Chatbot shows the pre-generated prompt incorporating music features, including extracted beat times, mood, and genre. After the Chatbot generates the raw trajectory, the user is able to pass it through our safety filter (AMSwarm), and simulate the filtered trajectory. Re-prompting is available at this step for the user to modify the trajectory using high-level language inputs. Finally, the deployment button will establish communication with the drones and execute the planned trajectories on the swarm.}
    \label{fig:appendix-gui}
\end{figure}

We use GPT-3.5 as the LLM to generate a time series of waypoints to choreograph the swarm of drones. To ensure an effective and generalizable prompt, we iteratively tune the initial prompt inspired by techniques from previous works like \cite{vemprala_chatgpt_2023} and based on our empirical experience from the performance of different prompts. The prompt is constructed with three core ideas: creativity, safety, and generalization. Fig.~\ref{fig:appendix-gui} shows the GUI for Swarm-GPT for prompting and re-prompting. It consists of a repository of songs for users to select, a Chatbot to show the pre-generated prompts and user reprompts, and an accurate simulation for the swarm trajectories. 

To make the choreography generated by GPT artistic, synchronized to music, and creative, we first assign GPT the role of an expert choreographer for a swarm of small drones. We use beat times to guide the music synchronization to mimic the human choreography process. The extracted timestamps of the beats of the song (gathered using Librosa) are fed into the prompt, where GPT is instructed to change the formation of the drone swarm at every beat by generating unique waypoints for each drone at each beat timestamp. We also prompt GPT to interpret the mood and tone of the well-recognized songs so that it can generate different types of choreography for different music genres. Keywords such as ``harmonic'', ``symmetric'', ``artistic'', and ``synchronized'' are also included in the prompt to drive GPT to generate aesthetic choreographs. 

The second aspect of the prompt emphasizes safety. Although we are using AMSwarm as an additional safety layer, we still prompt GPT to handle some rudimentary collision avoidance and constraint violations. We input the following constraints into the prompt: physical limitations of the flight volume, maximum drone velocity and acceleration, and explicitly instruct GPT to avoid having two waypoints that are too close to each other at the same timestamp.

To make the interface more robust and easy to use for non-experts, we make the prompt generalizable by feeding in the initial positions of the drones detected from the configuration file and specifying the output format so that it could be seamlessly processed by the next module in the pipeline.

Finally, we instruct GPT to format its output as a series of waypoints for each drone. This waypoint format provides GPT full autonomy to generate a waypoint-based choreography creatively since it is not restricted to a specific library of motion primitives or maneuvers. This format is also easy for GPT to interact with and does not require domain-specific robotics knowledge that GPT may not be trained on or augmented with.

GPT is asked to represent its response as a string that can be easily parsed into a JSON object. This was found to be the simple and reliable way to extract numerical data from GPT. The outputted string is verified for correct formatting, with superfluous text discarded and parsed automatically into JSON format. The JSON object is then converted to an array of waypoints sorted by timestamp.


\subsection{AMSwarm-Based Safety Filter Design}
\label{methodology:collision-avoidance}

Waypoints generated by the LLM are not always feasible and require preliminary processing and checks before being processed by our formal collision-avoidance algorithm or deployed onto the quadrotors. To ensure that the waypoints generated by the LLM are within the limits of the permissible flying region, the raw waypoints are normalized to fit the volume whenever they violate positional constraints.

The output is then checked to ensure no simultaneously occurring waypoints are within a certain threshold distance of each other. If they are, these waypoints are pulled apart by a pre-selected offset, thus eliminating the collision. Without this step, the formal collision avoidance algorithm outlined in the next paragraph would begin in a state of collision and would not converge on a solution.

The formal collision avoidance features an alternating minimization approach known as AMSwarm \cite{adajania_amswarm:_2023} to process the LLM-generated waypoints. It receives the pre-processed waypoints and interpolates between them in a way that ensures the satisfaction of physical constraints while minimizing inter-agent collisions. Unlike other collision-avoidance methods that struggle with approximations and linearizations, this approach reformulates constraints in a polar form and applies an alternating minimization (AM) algorithm to solve the resultant optimization problem, all while maintaining the quadratic program (QP) structure \cite{adajania_amswarm:_2023}.

For a single drone $i$ of the swarm size $N$, the initial and goal positions ($p_{i,o}$, $p_{i,g}$) are described with the three-dimensional position vector $p = [x,y,z]^T$ and are given by the LLM choreography. At each time step~$k$, each drone $i$ solves the following constrained optimization problem:

\begin{align}
    \min_{p_i}\hspace{1em} &\ \omega_g \sum_{k=K-\kappa}^{K-1} \left\Vert p_i[k] - p_{i,g}\right\Vert^2 + \omega_s \sum_{k=0}^{K-1} \left\Vert {p_i}^{(q)}[k]\right\Vert^2  \label{eq:min_p} \\ 
    \text{subject to} \ \hspace{1em}&{p_i}^{(q)}[0] = {p}^{(q)}_{i,a}, \ \forall q = \{0,1,2\},  \label{eq:p_init} \\
    & \underline{p} \preceq p_i[k] \preceq \overline{p}, \ \forall k, \label{eq:p_bound_1} \\
    & \left\Vert \dot{p}_i[k]\right\Vert^2 \leq \overline{v}^2, \ \forall k, \label{eq:p_bound_2} \\
    & \underline{f}^2 \leq \left\Vert \ddot{p}_i[k] + g\right\Vert^2  \leq \overline{f}^2, \ \forall k, \label{eq:p_bound_3} \\
    & h_{ij}[k] = \left\Vert \Theta_{ij}^{-1}({p}_i[k] - \xi_j[k])\right\Vert^2 - 1 \geq 0, \ \forall k,j, \label{eq:col_avoid}
\end{align}

where $K$ denotes the planning horizon length, $\left\Vert \cdot \right\Vert$ the Euclidean norm, and the superscript $(q)$ the q-th time derivative of a variable. There are three components in the trajectory optimization problem, which are respectively described below: 

\textit{1) Cost Function:} The cost function in Eq.~\ref{eq:min_p} consists of two terms weighted by the constants $\omega_g$ and $\omega_s$. The former term, i.e., the goal cost, penalizes the deviation of the position of the drone from the specified goal position generated by the LLM over the last $\kappa < K$ steps in the prediction horizon. The latter, i.e., the smoothness cost, penalizes the $q$-th derivatives of the position trajectory.

\textit{2) Initialization and Feasibility Constraints:} The equality constraints in Eq.~\ref{eq:p_init} initialize the trajectory position and its higher derivatives for each drone. The conditions in Eq.~\ref{eq:p_bound_1}-\ref{eq:p_bound_3} enforce lower and upper boundaries on the position $(\underline{p}, \overline{p})$, velocity $(-\underline{v}, \overline{v})$, and acceleration $(\underline{f}, \overline{f})$. 

\textit{3) Collision Avoidance Constraints:} Eq.~\ref{eq:col_avoid} guarantees the avoidance of collision with either the $j$-th neighbouring drone (or an obstacle) at the position $\xi_j[k]$. The diagonal matrix $\Theta_{ij}$ consists of elements whose scalars characterize the lengths of the ellipsoidal envelopes for each axis around the neighbouring drone (or the obstacles). The vector $g$ denotes the gravitational acceleration. We can further extend the collision avoidance by using barrier function (BF) constraints to introduce different levels of safe behaviours~\cite{zeng2021safety}: 

\begin{equation}
    h_{ij}[k] - h_{ij}[k-1] \geq -\gamma \, h_{ij}[k-1], \ \forall k, j \label{eq:col_avoid_2},
\end{equation}
where $\gamma \in [0,1]$ controls how fast each drone is allowed to approach the constraint boundary given by $h_{ij} = 0$. We can set $\gamma = 1$ to restore the condition in Eq.~\ref{eq:col_avoid};  smaller values generally promote more gradual and conservative collision avoidance behaviour.


This approach involves iteratively optimizing trajectory parameters while ensuring that various constraints, such as collision avoidance, position constraints, and kinematic feasibility, are satisfied. The AM algorithm switches between optimizing different aspects of the trajectory, ensuring that each step adheres to the imposed constraints. This process produces trajectories that strike a balance between safety and efficiency.

This algorithm was selected due to its consistent and reliable performance in complex and dynamic environments as well as its scalability properties~\cite{adajania_amswarm:_2023}. Overall, the AMSwarm-based safety filter permits the choreography of larger drone swarms and future-proofs of our system to behave robustly within dense, interactive environments. The algorithm also interfaces well with our LLM outputs (i.e. waypoints). Other approaches require more information about the desired drone motions (for example, sinusoidal motion primitives), which can be difficult for language models to generate.

Our application of this algorithm features some minor changes. A new optimization problem is formulated for each inter-waypoint time window, and the alternating-minimization algorithm is re-applied for each one. The position control commands generated by optimization are then resampled at both the desired simulation and control frequencies prior to deployment.

The re-sampled position commands are simulated via gym-pybullet-drones \cite{panerati2021learning} as an on-demand, visual confirmation of collision-free motions.

\section{Additional Evaluation Results}
\label{appendix:results}
In this appendix, we present additional evaluation results for the proposed Swarm-GPT system. 

\begin{figure}[hbt!]
    \centering
    \input{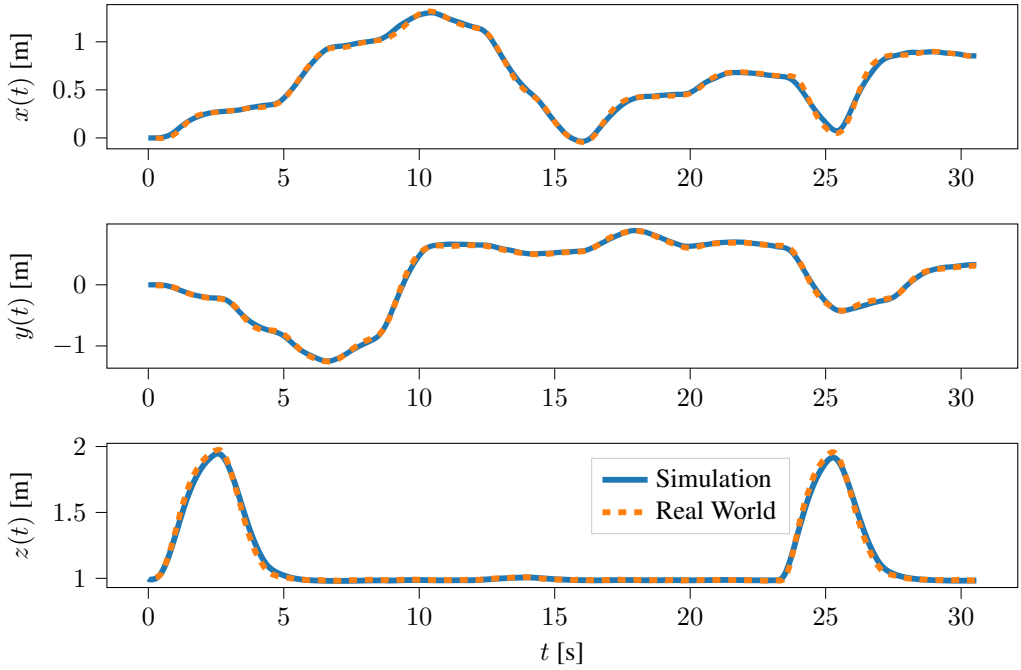}{}
    \caption{The simulated and actual position trajectories of a single drone. The simulated agent and the physical drone were given the same reference inputs. As can be seen from the plots, the actual positions of the drone closely match that of the simulated response.}
    \label{fig:appendix-rmse}
\end{figure}

\textbf{Assessment of Sim-to-Real Transfer.} Figure~\ref{fig:appendix-rmse} illustrates the close match between the simulated and real trajectory for one drone during a choreography. This corresponds to the low RMSE values we obtained in Fig.~\ref{fig:p}; this behaviour is generally observed across different drones and choreographies. 

\begin{figure}[!htb]
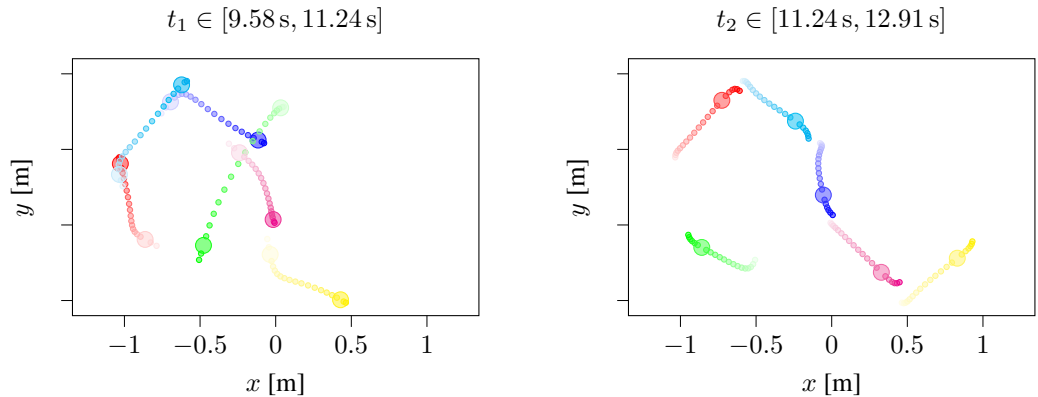

    \begin{minipage}{0.5\textwidth}
        \centering
        \input{figures/figure_6_2d_2}
    \end{minipage}~~
    \hspace{0.01\linewidth}
    \begin{minipage}{0.5\textwidth}
        \centering
        \input{figures/figure_6_2d_3}
    \end{minipage}
        \caption{A six-drone swarm 2D trajectory visualizations during one musical beat between 9.58 and 11.24 s \textit{(left)} and between 11.24 and 12.91 s \textit{(right)}. Similar to Fig.~\ref{fig:vis}, different colours correspond to different drones, the large dots represent beat times, and the colour gradient moves from lighter to more saturated as time progresses. These plots showcase the generative capability of the Swarm-GPT for creating interesting formations synchronized with music beats while complying with desired safety constraints.}
        \label{fig:6_2d_2}
\end{figure}

\textbf{Formation Synchronization to Music.} Figure~\ref{fig:6_2d_2} shows a synchronized motion among a swarm of 6 drones during two more beat intervals (in addition to that presented in Fig.~\ref{fig:vis}). These results demonstrate the capability of Swarm-GPT to generate creative and expressive drone formations in synchronization with the music features.

\end{document}